\pdfoutput=1

\documentclass[11pt]{article}

\usepackage{acl}

\usepackage{times}
\usepackage{latexsym}

\usepackage[T1]{fontenc}

\usepackage[utf8]{inputenc}

\usepackage{microtype}

\usepackage{natbib}  
\usepackage{caption} 
\usepackage{amsopn}
\usepackage{booktabs}
\usepackage{multirow}
\usepackage{etoolbox}
\usepackage{textpos} 
\usepackage{helvet}  
\usepackage{courier}  
\usepackage{graphicx} 
\usepackage{algorithm}
\usepackage{algorithmic}
\usepackage{CJKutf8}
\usepackage[utf8]{inputenc}
\usepackage{natbib}
\usepackage{xcolor}
\usepackage[algo2e]{algorithm2e}
\usepackage{array}
\usepackage{newfloat}
\usepackage{listings} 

\usepackage[namelimits]{amsmath} 
\usepackage{amssymb}             
\usepackage{amsfonts}            
\usepackage{mathrsfs}            
\usepackage{amsopn}
\usepackage{multicol}

\title{TransAug: Translate as Augmentation for Sentence Embeddings}

\author{Jue Wang}

\begin{document}
\maketitle
\begin{abstract}

While contrastive learning greatly advances the representation of sentence embeddings, it is still limited by the size of the existing sentence datasets. In this paper, we present TransAug (\textbf{Trans}late as \textbf{Aug}mentation), which provide the first exploration of utilizing translated sentence pairs as data augmentation for text, and introduce a two-stage paradigm to advances the state-of-the-art sentence embeddings. Instead of adopting an encoder trained in other languages setting, we first distill a Chinese encoder from a SimCSE encoder (pretrained in English), so that their embeddings are close in semantic space, which can be regraded as implicit data augmentation. Then, we only update the English encoder via cross-lingual contrastive learning and frozen the distilled Chinese encoder. Our approach achieves a new state-of-art on standard semantic textual similarity (STS), outperforming both SimCSE and Sentence-T5, and the best performance in corresponding tracks on transfer tasks.

\end{abstract}

\section{Introduction}

It has been a fundamental problem in natural language processing to learn sentence embeddings that provide compact semantic representations~\cite{le2014distributed,gao2021simcse,ni2021sentence,reimers2019sentence,wang2021efficientclip, wang2022eclip, gao2025foldtoken,gao2021distilcse,gao2024uniif,tan2023revisiting,tan2025ustep}. Recently, contrastive learning (CL) which aims to learn effective representation by pulling semantically close neighbors together and separate non-neighbors, has widely attracted attention for building universal representations. It is noteworthy that benefit from powerful contrastive learning framework, scaling up the size of dataset greatly improve robustness and generalization of representations as suggested by some previous works~\cite{radford2021learning,chen2020improved,jia2021scaling,wang2021efficientclip}.



\begin{figure*}[t]
\centering
\includegraphics[scale=0.20]{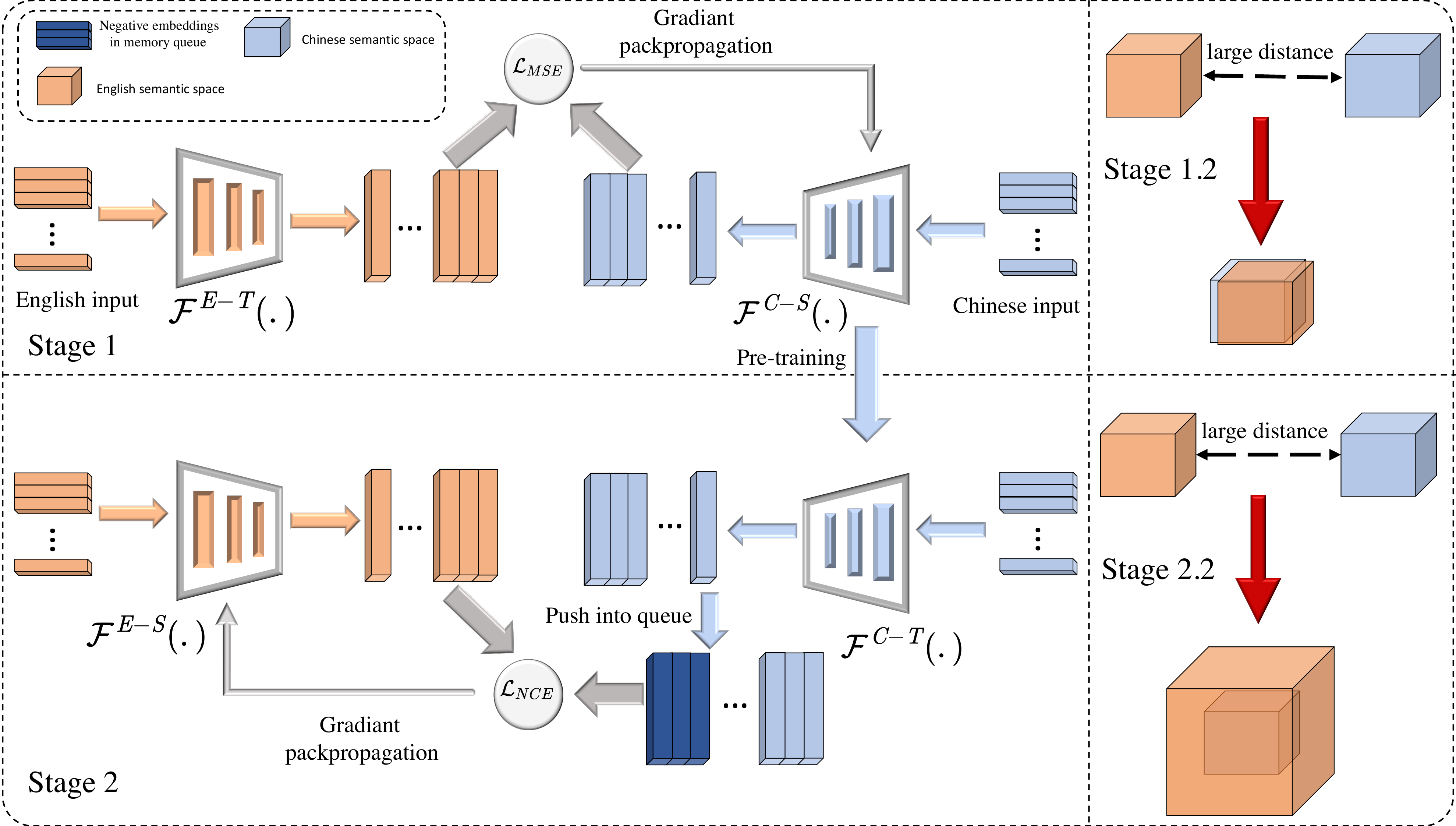}
\caption{\textbf{The pipeline of two-stage TransAug.} Stage 1 and stage 2 describe the distillation and contrastive learning process, respectively. $F^{E-T}$, $F^{C-S}$, $F^{E-S}$, $F^{C-T}$ represent English-teacher, Chinese-student, English-student, Chinese-teacher, respectively. Stage 1.2 and stage 2.2 represent the target of optimization in different stage, the goal in stage 1 is to minimize the distance, while the goal in stage 2 is to discriminate.}
\label{fig:pipeline}
\end{figure*}

SimCSE~\cite{gao2021simcse} demonstrates that a contrastive objective can be extremely effective when coupled with pre-trained language models. However, the generality and capability of the language model is strictly limited by the size of parallel sentence pairs (less than 1 million). To alleviate this issue, it is sensible and practical to construct a comparably large-scale paired sentence dataset through translation, inspired by previous works~\cite{pan2021contrastive,feng2020language,yang2019improving} in multilingual filed, yet there is no efficient way to utilize the translated pairs for sentence representation learning.

In this paper, we provide the first exploration
of using translated sentence pairs as data augmentation (TransAug) for text, and introduce a two-stage
paradigm to learn superior sentence embeddings. To construct positive embedding pairs for contrastive learning, the most naive idea is to employ two independent encoders trained on different language datasets (Chinese and English in our case) to produce sentence embeddings given the translated pair as input, or adopt a single encoder which is able to accept multi-language input. However, due to the distribution deviation of different language inputs, the generated two embeddings usually can not smoothly lie in the same representation space, 
which degrades the power of contrastive learning. Thus, instead of directly adopting an existed SimCSE (trained in Chinese) model, we first conduct multilingual semantic distillation (MSD) to obtain a Chinese encoder from a pretrained SimCSE model (trained in English), so that their embeddings are close in semantic space and can be regarded as implicit data augmentation. In stage two, we propose a Cross-lingual contrastive method and a multilingual teacher-student contrastive architecture, where the distilled Chinese encoder (as teacher) is frozen and supervise the English encoder (as student) through contrastive loss~\cite{hadsell2006dimensionality}. Specifically, the student encoder produces query embeddings and the teacher encoder generates key embeddings and negative embeddings, the objective is to distinguish whether the query embeddings match the corresponding key embeddings. The pipeline is illustrated in Figure \ref{fig:pipeline}.

To better validate the predominant performance of
TransAug, we conduct a comprehensive evaluation protocol following the same setting as SimCSE on seven standard semantic textual similarity (STS) tasks~\cite{agirre2012sem,agirre2013sem,agirre2014semeval,agirre2015semeval,agirre2016semeval,cer2017semeval,marelli2014sick} and seven transfer tasks~\cite{conneau2018senteval}. TransAug achieves a new state-of-art on STS tasks, outperforming SimCSE and Sentence-T5~\cite{ni2021sentence} by margin, and also achieves state-of-art performance in corresponding tracks on transfer tasks evaluated by SentEval~\cite{conneau2018senteval}. On the average score of STS tasks, our pre-trained TranAug-BERT${base}$ with or without fine-tuning surpass SimBert$_{base}$ by 3.58\% and 2.65\% respectively, and TranAug-RoBERTa$_{large}$ achieves 85.60 on average. Surprisingly, TranAug-bert$_{base}$ with fine-tuning achieves better results than Sentence-T5 (11B) with only 1\% parameters in comparison.

We summarize our contributions as below:

1. We provide the first exploration of using translated sentence
pairs as data augmentation for text.

2. A two-stage paradigm is introduced to utilize translated sentence pairs and improve the representation of sentence embeddings.

3. Our approach achieves a new state-of-the-art on standard semantic textual similarity (STS), and the best performance in corresponding tracks on transfer
tasks evaluated by SentEval.

\section{Related Work}

\subsection{Universal Sentence Representation} 


Sentence representation is a well-studied area with many proposed methods~\cite{mikolov2013efficient, pennington2014glove,le2014distributed}. With the progress of pre-training, pre-training objectives like BERT~\cite{devlin2018BERT}, RoBERTa~\cite{liu2019roberta} are utilized to generate the sentence embeddings. To derive sentence embeddings from BERT, Sentence-BERT~\cite{reimers2019sentence} use siamese and triplet network structures to derive semantically meaningful sentence embeddings that can be compared using cosine-similarity. SimCSE~\cite{gao2021simcse} introduce a simple contrastive learning framework, which greatly improves state-of-the-art sentence embeddings on semantic textual similarity tasks both on unsupervised and supervised tracks. Sentence-T5~\cite{ni2021sentence} investigates producing sentence embeddings from the pre-trained T5~\cite{raffel2019exploring} models and fine-tune them on downstream datasets that achieve the leading results in sentence embedding benchmark datasets.

\subsection{Multilingual Learning}

Multilingual learning has attracted increasing interests from the community. Parallel translation datasets have been widely leveraged for Neural Machine Translation (NMT), Semantic Retrieval (SR), Bitext Retrieval (BR) and Retrieval Question Answering (ReQA), etc. Multilingual Universal Sentence Encoder~\cite{yang2019multilingual} conduct a multitask trained dual encoder to bridge 16 different languages, and achieves competitive results on SR, BR, ReQA tasks. LaBSE~\cite{feng2020language} adopt a  dual encoder with additive margin softmax combined masked language model (MLM) and translation language model (TLM) to improve multilingual sentence embeddings. mRASP2~\cite{pan2021contrastive} hypothesis that a universal cross-lingual representation leads to better multilingual translation performance. They regard a corresponding pair as a positive sample, and other in-batch samples including a variety of languages as negative samples, to establish a contrastive learning process. In this way, multiple languages representations are smoothly embedded into a close semantic space. Different from previous works that focus on embedding text from multiple languages into a close semantic space, we propose to utilize translation datasets as data augmentation or amplification for learning robust universal sentence embedding.

\section{Methods}
\label{sec:length}

In this section, we first compare two practical strategies for translated sentence pairs, then illustrate the two-stage paradigm of our proposed TransAug and show the necessity of each stage. The pipeline of TransAug is shown in Figure \ref{fig:pipeline}.

\begin{figure}[h]
\begin{center}
\includegraphics[width=0.5\textwidth]{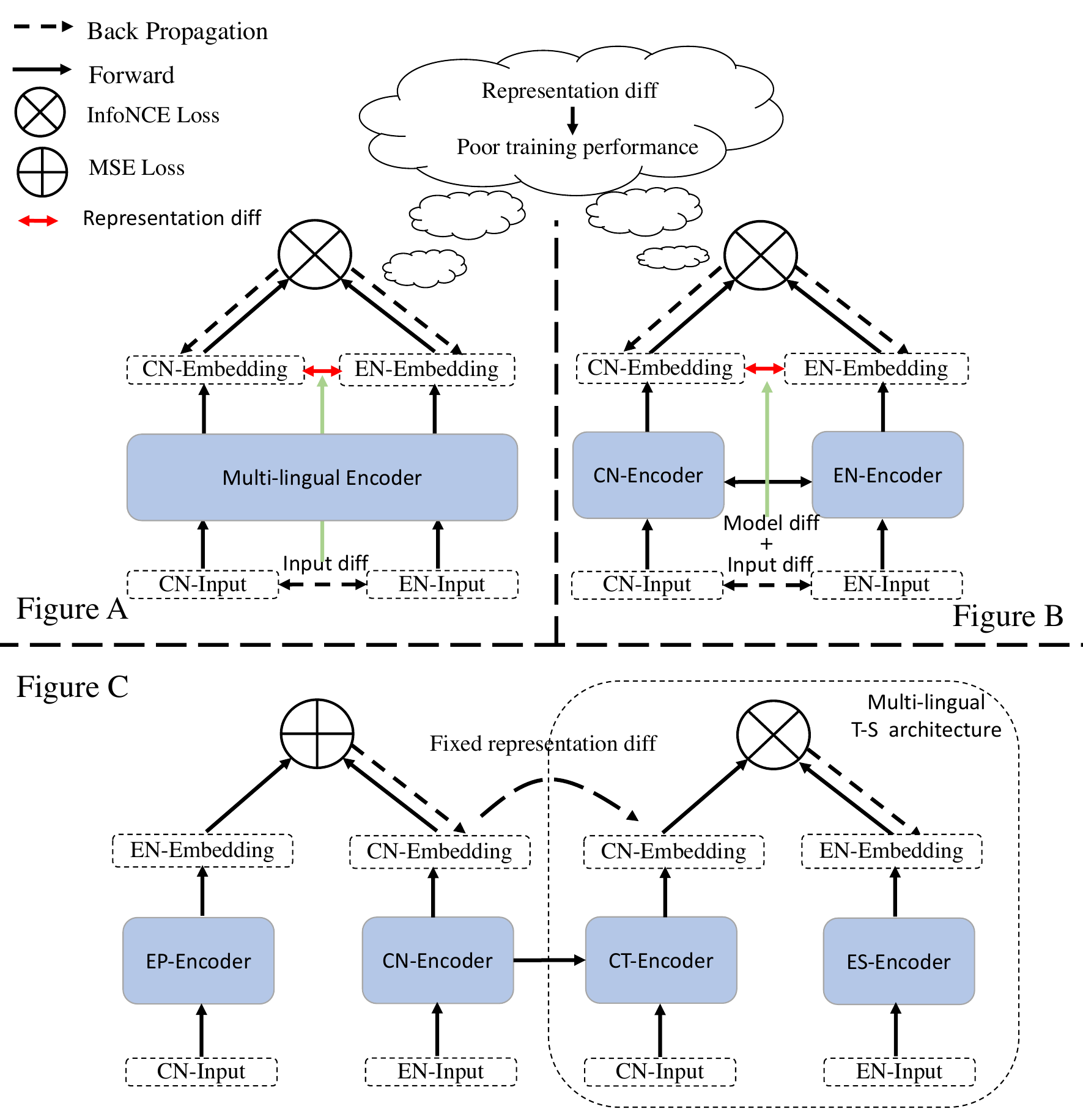}
\vspace{-6mm}
\caption{\textbf{Comparison of three strategies for translated sentence pairs.} Figure A, Figure B and Figure C represent single multilingual encoder, regular dual encoder and our TransAug respectively.}

\label{fig:comparison}
\end{center}
\vspace{-6mm}
\end{figure}

\subsection{Preliminary}
\label{Preliminary}

In this subsection, we briefly introduce two preliminaries for utilizing paired, which have been commonly used in contrastive learning approaches.

\textbf{Multilingual Single Encoder~\cite{yang2019multilingual,pan2021contrastive}} embeds sentence from different languages into a single semantic space using a unified encoder, based on the hypothesis that a universal cross-language learning leads to better sentence representation. Its architecture is illustrated in A, Figure \ref{fig:comparison}.

\textbf{Dual Encoder~\cite{he2020momentum,radford2021learning,ni2021sentence}},  also known as two-tower, models the paired data with two separate encoders, and project the representation of paired input into the same embedding space through jointly training. Its architecture is illustrated in B, Figure \ref{fig:comparison}.

However, these methods are not designed to learn universal representation of sentence and lead to poor generalization. Thus, to make up this gap, we slightly modify the dual encoder architecture and propose a two-stage paradigm (TransAug) to advance the representation of sentence. We introduce the two-stage in Section \ref{MSD} and Section \ref{ccl} respectively, and conduct comprehensive experiments to verify the effectiveness of TransAug compared with two preliminaries in Section \ref{MTA}. The simplified comparison is shown in Figure \ref{fig:comparison}.

\subsection{Multilingual Semantic Distillation}
\label{MSD}



\begin{figure}[h]
\begin{center}
\includegraphics[width=0.4\textwidth]{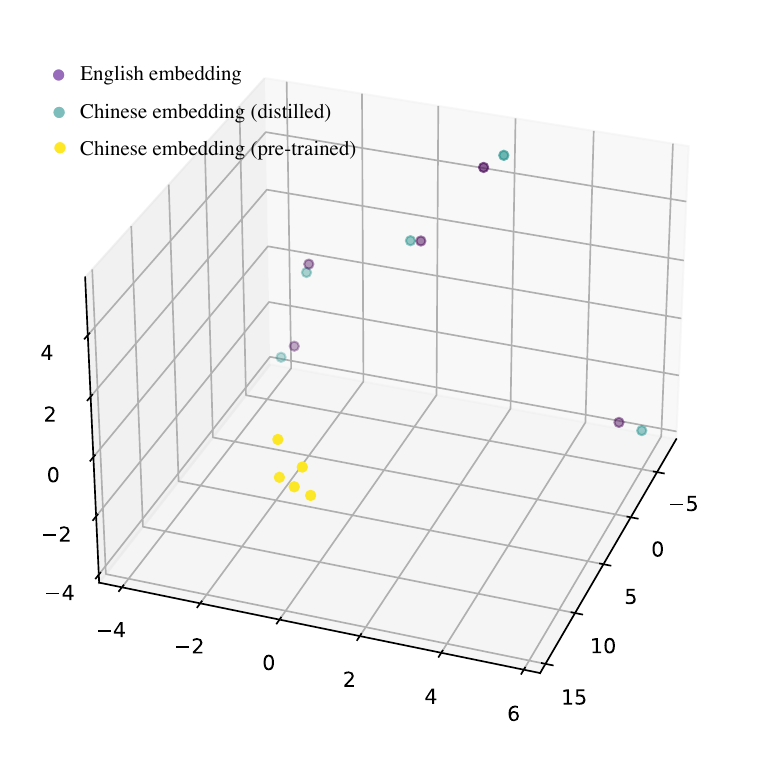}
\vspace{-6mm}
\caption{\textbf{Embedding similarity in semantic space.} The English embedding is generated by SimCSE (trained in English), the Chinese embeddings are generated by SimCSE (trained in Chinese) and our distilled model. As shown, the distilled Chinese embeddings are much closer to English embeddings.}

\label{fig:distribution_MSD}
\end{center}
\vspace{-4mm}
\end{figure}

In the first stage, we conduct multilingual semantic distillation (MSD) to obtain a Chinese sentence encoder that has similar semantic space as the English sentence encoder. Specifically, we freeze the pre-trained SimCSE model (trained in English) and use its embeddings to supervise a BERT or RoBERTa (pre-trained in Chinese), and minimize the L2 distance between the English embeddings and Chinese embedding using an MSE loss. 

\textbf{Why not use a pre-trained encoder?} To encode the translated sentence pair, the most direct way is adopting an existed pre-trained encoder (trained in Chinese). However, as the distribution deviation of language datasets, the generated two embeddings usually do not lie in the similar representation space, which degrades the power of contrastive learning. Thus, instead of directly adopting an existed SimCSE
(trained in Chinese) model, we first conduct multilingual semantic distillation to obtain a Chinese encoder from a pretrained SimCSE model (trained in English), so that their embeddings are close in semantic space and can be regarded as implicit data augmentation. To validate the hypothesis that distilled Chinese embedding is closer to English embedding, we randomly sample 5 translated pairs (Chinese-English sentences) from the WMT\footnote{http://www.statmt.org/wmt20/} validation set, and visualize the generated embeddings through PCA~\cite{shlens2014tutorial} method. As shown in Figure \ref{fig:distribution_MSD}, the distilled Chinese embedding is indeed closer to the corresponding English embedding than its counterpart from a pre-trained Chinese encoder. To better show the effectiveness of multilingual semantic distillation, we also conduct an ablation study in Section \ref{exp of msd} to confirm that the distilled Chinese encoder is superior to pre-trained Chinese encoder.

\subsection{Cross-Lingual Contrastive Learning}
\label{ccl}

After we obtain a distilled Chinese encoder that generate embeddings closing to English sentence encoder from stage one, the next key step is to apply contrastive objectives on translation datasets. Different from previous works that adopt a single~\cite{gao2021simcse,pan2021contrastive} or dual ~\cite{he2020momentum,radford2021learning,ni2021sentence} encoder as backbone, TransAug introduces a new-fashioned multilingual teacher-student architecture to conduct contrastive learning effectively. Different from SimCSE\cite{gao2021simcse} that apply dropout as augmentation method, in our case, we claim that distillation in stage one can be regarded as a kind of implicit data augmentation, where a translated sentence pairs and their embeddings generated by our teacher-student architecture establish positive samples in contrastive learning framework.

\begin{figure}[h]
\begin{center}
\includegraphics[width=0.4\textwidth]{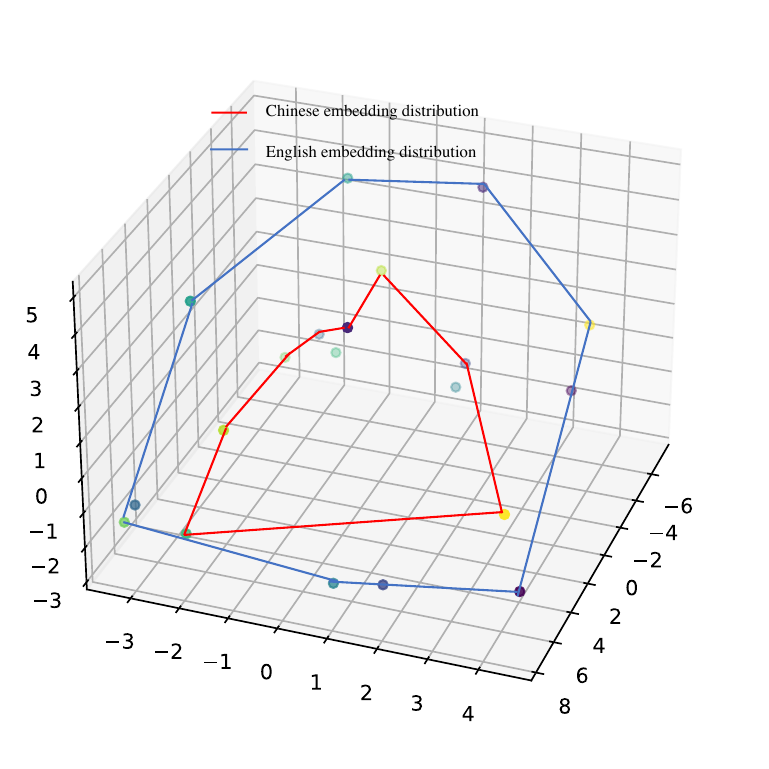}
\vspace{-6mm}
\caption{\textbf{Embedding distribution of the teacher and student model in CCL.} The dots in the same color are the representations of the corresponding pair. The dots connected by the solid red line are in Chinese, and the solid blue line is in English.}

\label{fig:distribution_CCL}
\end{center}
\vspace{-4mm}
\end{figure}

We assign the distilled Chinese encoder from stage one as teacher model and a pre-trained BERT or RoBERTa model as student model. Then, we freeze the teacher model and only use its consistent embeddings to build a large memory queue, and produce the key embeddings. For student model, it supplies the query embedding to match the key embedding via contrastive learning framework.

Notably, we notice that using the frozen robust embeddings from teacher model as the contrastive labels to separate the student model's embeddings greatly encourages the training efficiency and makes embedding from student model more discriminative. As visualized in Figure \ref{fig:distribution_CCL}, where we randomly sample 10 pairs from WMT evaluation set and visualize the embedding distribution, the semantic space of the student model is larger than its teacher counterpart. We provide more analysis in Section \ref{MTA}.



\section{Experiments}

We first introduce the datasets adopted in our work, and illustrate details of each module in our proposed framework. Then, we conduct experiments on 7 semantic textual similarity (STS) tasks following previous work~\cite{gao2021simcse,ni2021sentence}. We also evaluate 7 transfer learning tasks and provide detailed results. Finally, we do ablation studies to validate the effectiveness of proposed modules.


\subsection{Datasets}

For the two-stage training process, we use two datasets in our work: WMT dataset and a large-scale dataset collecting from the internet.

\subsubsection{WMT Dataset}


This is a common-used machine translation dataset composed of a collection of various sources. We translate the original sentence in English to Chinese. The corpus file has 19,442,200 Chinese-English parallel sentence pairs.

\subsubsection{Source-mixed Dataset}


To further scale up the size of the training dataset, we extra collect open-sourced translation datasets from the internet on the top of WMT dataset, including AIC~\cite{wu2017ai}, translation2019zh~\cite{bright_xu_2019_3402023}, etc. Finally, we establish a larger-scale dataset including 67,307,798 Chinese-English parallel pairs.

\subsection{Training Details}

We elaborate the training details of Multilingual Semantic Distillation (MSD) and Cross-Lingual Contrastive Learning (CCL), respectively. All experiments are conducted on 8 NVIDIA V100 GPUs.


\subsubsection{Multilingual Semantic Distillation}


In the stage one of TransAug, the main goal is to obtain a Chinese encoder that generate embeddings closed to the original English embeddings in semantic space. Specifically, we adopt the pre-trained SimCSE-RoBERTa$_{large}$ model as the English-teacher encoder, and establish a RoBERTa$_{large}$ model as the Chinese-student encoder. We set learning rate to 0.00005, batch size to 160, dropout to 0.1, and the input sentence length to 50. In addition, a cosine learning rate scheduler is applied for maintaining the consistency of training. We freeze the teacher encoder and only update the student model as regular multilingual semantic distillation setting, which minimize the distance between English and Chinese embeddings. The student model is trained for 2 epochs with source-mixed dataset.


\begin{table*}[h]
    \small
    \centering
    \resizebox{\linewidth}{!}{%
    \begin{tabular}{l|c|rrrrrrrr}
\textbf{Model} & \textbf{Fine-tune data} & \textbf{STS12}    & \textbf{STS13}    & \textbf{STS14}  & \textbf{STS15}  & \textbf{STS16}   & \textbf{STSb}  & \textbf{SICK-R}  & \textbf{Avg}   \\
 \hline
 \rule{-2pt}{10pt}
 SBERT$_{base}$  & NLI & 70.97 & 76.53 & 73.19 & 79.09 & 74.30 & 77.03 & 72.91 & 74.89 \\
 SBERT$_{base}$-flow  & NLI & 69.78 & 77.27 & 74.35 & 82.01 & 77.46 & 79.12 & 76.21 & 76.60 \\
 SBERT$_{base}$-whitening & NLI & 69.65 & 77.57 & 74.66 & 82.27 & 78.39 & 79.52 & 76.91 & 77.00 \\
 CT-SBERT$_{base}$ & NLI & 74.84 & 83.20 & 78.07 & 83.84 & 77.93 & 81.46 & 76.42 & 79.39 \\
 SimCSE-BERT$_{base}$ & NLI & 75.30 & 84.67 & 80.19 & 85.40 & 80.82 & 84.25 & 80.39 & 81.57  \\
 TransAug-BERT$_{base}$(WMT) & N/A & 80.13 & 86.80 & 83.22 & 88.72 & 82.42 & 86.73 & 81.15 & 84.17  \\
 TransAug-BERT$_{base}$(SMD) & N/A & 79.21 & 87.84 & 83.24 & 88.64 & 82.42 & 86.87 & 81.31 & 84.22  \\
 TransAug-BERT$_{base}$(WMT) & NLI & \textbf{81.08} & 88.19 & \textbf{84.07} & 88.28 & 84.48 & 87.14 & 81.36 & 84.94  \\
 TransAug-BERT$_{base}$(SMD) & NLI & 80.26 & \textbf{88.70} & 84.05 & \textbf{88.62} & \textbf{84.57} & \textbf{87.95} & \textbf{81.87} & \textbf{85.15}  \\
 \hline
 \rule{-2pt}{10pt}
 SBERT$_{large}$ & NLI & 72.27 & 78.46 & 74.90 & 80.90 & 76.25 & 79.23 & 73.75 & 76.55 \\
 SimCSE-BERT$_{large}$ & NLI & 75.78 & 86.33 & 80.44 & 86.60 & 80.86 & 84.87 & 81.14 & 82.21  \\
 TransAug-BERT$_{large}$(WMT) & N/A & 78.41 & 87.23 & 83.21 & 89.08 & 82.97 & 87.00 & 81.65 & 84.22  \\
 TransAug-BERT$_{large}$(SMD) & N/A & 79.18 & 87.75 & 82.85 & 88.53 & 82.60 & 86.85 & 81.51 & 84.18  \\
 TransAug-BERT$_{large}$(WMT) & NLI & 80.86 & 88.93 & 84.01 & 88.81 & 84.71 & 87.96 & 81.03 & 85.19  \\
 TransAug-BERT$_{large}$(SMD) & NLI & \textbf{80.86} & \textbf{89.47} & \textbf{84.35} & \textbf{88.97} & \textbf{85.04} & \textbf{88.58} & \textbf{81.63} & \textbf{85.56}  \\
 \hline
 \rule{-2pt}{10pt}
 SRoBERTa$_{large}$-whitening & NLI & 74.53 & 77.00 & 73.18 & 81.85 & 76.82 & 79.10 & 74.29 & 76.68 \\
 SimCSE-RoBERTa$_{large}$ & NLI & 77.46 & 87.27 & 82.36 & 86.66 & 83.93 & 86.70 & 81.95 & 83.76  \\
 TransAug-RoBERTa$_{large}$(WMT) & N/A & 79.19 & 87.52 & 83.67 & 88.92 & 83.03 & 87.13 & 81.51 & 84.42  \\
 TransAug-RoBERTa$_{large}$(SMD) & N/A & 79.42 & 88.12 & 83.71 & 88.95 & 83.37 & 87.20 & 81.76 & 84.65  \\
 TransAug-BRoBERTa$_{large}$(WMT) & NLI & \textbf{80.73} & 88.93 & 84.52 & \textbf{88.80} & 84.44 & 88.29 & \textbf{81.99} & 85.39  \\
 TransAug-RoBERTa$_{large}$(SMD) & NLI & 80.39 & \textbf{89.62} & \textbf{84.76} & 88.67 & \textbf{85.06} & \textbf{88.58} & \textbf{82.15} & \textbf{85.60}  \\
 \hline
 \rule{-2pt}{10pt}
 ST5-Enc mean (11B) & NLI & 77.42 & 87.50 & 82.51 & 87.47 & 84.88 & 85.61 & 80.77 & 83.74 \\
 ST5-EncDec first (11B) & NLI & 80.11 & 88.78 & 84.33 & 88.36 & 85.55 & 86.82 & 80.60 & 84.94  \\
 TransAug-BERT$_{base}$(SMD) & NLI & 80.26 & 88.70 & 84.05 & 88.62 & 84.57 & 87.95 & 81.87 & 85.15  \\
 TransAug-BERT$_{large}$(SMD) & NLI & \textbf{80.86} & 89.47 & 84.35 & \textbf{88.97} & 85.04 & 88.58 & 81.63 & 85.56  \\
 TransAug-RoBERTa$_{large}$(SMD) & NLI & 80.39 & \textbf{89.62} & \textbf{84.76} & 88.67 & \textbf{85.06} & \textbf{88.58} & \textbf{82.15} & \textbf{85.60}  \\
 \hline
    \end{tabular}
    }
    \caption{\textbf{Comparison with previous state-of-the-art works in STS task.} All results are from ~\citealp{gao2021simcse,ni2021sentence,reimers2019sentence}; WMT and SMD represent the model is trained on WMT dataset and source-mixed dataset, respectively.}
    \label{tab:sts}
\end{table*}

\subsubsection{Cross-Lingual Contrastive Learning}

After obtain a distilled Chinese encoder from stage one, the next step is to conduct efficient contrastive learning for utilizing cross-lingual embeddings. To be more specific, we froze the parameters of the distilled encoder, and align the same training setting as SimCSE. We evaluate every 250 training steps on the dev set of STS-B and keep the best checkpoint for the final evaluation on test sets.

We also provide comprehensive analysis of hyperparameters on cross-lingual contrastive learning, including the size of negative sample queue, learning rate and  batch size. All experiments are conducted on STS-B development set.

\textbf{Size of Memory Queue.} The negative sample queue is a critical component in the contrastive learning framework. We analyze the effect of queue size for different student backbones (BERT$_{base}$ and RoBERTa$_{large}$) on cross-lingual learning process.

\renewcommand{\arraystretch}{1.2}
\begin{table}[H]
  \centering
  \begin{tabular}{c||c|c|c|c}
    \hline
     Queue size & 1024 & 4096 & 10T & 50T \\
    \hline
    BERT$_{base}$  &87.82 &\textbf{88.08}  &87.79 & 87.92
 \\
    \hline
  \end{tabular}
  \vspace{-2mm}
    \caption{Effect of the queue size on BERT$_{base}$. T is the abbreviation for \textbf{T}housands.}
  \label{queue size of bert}
  \vspace{-2mm}
\end{table}

\renewcommand{\arraystretch}{1.2}
\begin{table}[H]
  \centering
  \begin{tabular}{c||c|c|c|c}
    \hline
     Queue size & 10T & 50T & 200T & 300T\\
    \hline
    RoBERTa$_{large}$  &87.95 &88.04  &\textbf{88.36} & 87.36
 \\
    \hline
  \end{tabular}
  \vspace{-2mm}
    \caption{Effect of the queue size on RoBERTa$_{large}$. T is the abbreviation for \textbf{T}housands.}
  \label{queue size of roberta}
  \vspace{-2mm}
\end{table}

As shown in Table \ref{queue size of bert} and Table \ref{queue size of roberta}, BERT$_{base}$ achieves the best result with a small queue size, while RoBERTa$_{large}$ requires a large queue size for better performance.

\textbf{Effect of Learning Rate.} We perform grid searching for finding a suitable learning rate both for BERT$_{base}$ and RoBERTa$_{large}$. The results are reported in \ref{lr bert} and \ref{lr roberta} repectively.

\renewcommand{\arraystretch}{1.2}
\begin{table}[H]
  \centering
  \begin{tabular}{c||c|c|c|c}
    \hline
     LR & 5e-5 & 1e-4 & 2e-4 & 5e-4 \\
    \hline
    BERT$_{base}$  &86.16 &87.95  &\textbf{88.08} & 84.68 \\
    \hline
  \end{tabular}
  \vspace{-2mm}
    \caption{Effect of the learning rate on BERT$_{base}$.}
  \label{lr bert}
  \vspace{-2mm}
\end{table}

\renewcommand{\arraystretch}{1.2}
\begin{table}[H]
  \centering
  \begin{tabular}{c||c|c|c|c}
    \hline
     LR & 1e-5 & 2e-5 & 5e-5 & 1e-4\\
    \hline
    RoBERTa$_{large}$ &86.47 &86.86  &\textbf{88.36} & 87.76
 \\
    \hline
  \end{tabular}
  \vspace{-2mm}
    \caption{Effect of the learning rate on RoBERTa$_{large}$.}
  \label{lr roberta}
  \vspace{-2mm}
\end{table}

\textbf{Effect of Batch Size.} As shown in Table \ref{batch size bert}, we set the batch size of BERT$_{base}$ to 400 for the best result. Restricted by the computing resource, 160 is the largest batch size we can set for RoBERTa$_{large}$.

\renewcommand{\arraystretch}{1.2}
\begin{table}[H]
  \centering
  \begin{tabular}{c||c|c|c|c}
    \hline
     Batch size & 128 & 256 & 400 & 480 \\
    \hline
    BERT-base  &85.33 &87.87  &\textbf{88.08} & 88.00 \\
    \hline
  \end{tabular}
  \vspace{-2mm}
    \caption{Effect of batch size on BERT$_{base}$}
  \label{batch size bert}
  \vspace{-2mm}
\end{table}

\renewcommand{\arraystretch}{1.2}
\begin{table}[H]
  \centering
  \begin{tabular}{c||c|c|c|c}
    \hline
     Batch size & 64 & 128 & 160 & / \\
    \hline
    RoBERTa$_{large}$  & 87.75 & 87.82 & \textbf{88.36} & /
 \\
    \hline
  \end{tabular}
  \vspace{-2mm}
    \caption{Effect of batch size on RoBERTa$_{large}$.}
  \label{batch size roberta}
  \vspace{-2mm}
\end{table}

\textbf{Effect of Temperature.} Temperature is a crucial factor which impact the coverage of training and the model's performance in contrastive learning. We evaluate a number of temperatures recommended by previous works~\cite{gao2021simcse,ni2021sentence,radford2021learning}, including 0.05, 0.01, learnable temperature 1 (a learnable parameter in training). As shown in Table~\ref{temperature}, a learnable temperature 1 works best.

\renewcommand{\arraystretch}{1.2}
\begin{table}[H]
  \centering
  \begin{tabular}{c||c|c|c}
    \hline
     Temperature & 0.01 & 0.05 & L1 \\
    \hline
    BERT$_{base}$  &82.21 &86.80  &\textbf{88.08}  \\
    \hline
  \end{tabular}
  \vspace{-2mm}
    \caption{\textbf{Effect of the temperature.} L1 represents the learnable temperature 1.}
  \label{temperature}
  \vspace{-2mm}
\end{table}

For BERT$_{base}$, the learning rate is 0.0002, batch size is 400, queue size is 4096, and the dropout is defaulted set as 0.1. We leverage the cosine learning rate scheduler to adjust the learning rate dynamically. In the term of RoBERTa$_{large}$, we set the learning rate to 0.00005, batch size to 160, queue size to 200,000, all other hyperparameters keep the same as BERT$_{base}$.

\subsubsection{Finetune on NLI Dataset}
We investigate whether more training data are additive for better sentence representations by fine-tuning on NLI dataset (SNLI~\cite{bowman2015large} and MNLI~\cite{williams2017broad}). The NLI dataset contains 275,602 samples and each sample is consisted of a query sentence, a positive sentence, and a hard negative sentence. Following a similar training setting of SimCSE, we set the learning rate to 0.00001, batch size to 128, dropout to 0.1, temperature to 0.05 and input length to 50 for small models (BERT$_{base}$ and RoBERT$_{base}$). While for large models (BERT$_{large}$ and RoBERTa$_{large}$), we set learning rate to 0.00001, batch size to 96.

\begin{table*}[h]
    \small
    \centering
    \begin{tabular}{l|rrrrrrrr}
\textbf{Model}  & \textbf{MR}    & \textbf{CR}    & \textbf{SUBJ}  & \textbf{MPQA}  & \textbf{SST}   & \textbf{TREC}  & \textbf{MRPC}  & \textbf{Avg}   \\
 \hline
 \rule{-2pt}{10pt}
         InferSent-GloVe & 81.57 & 86.54 & 92.50 & 90.38 & 84.18 & 88.20 & 75.77 & 85.59\\
        Universal Sentence Encoder & 80.09 & 85.19 & 93.98 & 86.70 & 86.38 & 93.20 & 70.14 & 85.10  \\
        \midrule
        SBERT$_{base}$ & 83.64&	89.43&	94.39&	89.86 &	88.96&	89.60&	76.00&	87.41\\
        SimCSE-BERT$_{base}$  & 82.69&	89.25&	\textbf{94.81}&	89.59&	87.31&	88.40&	73.51&	86.51\\
        TransAug-BERT$_{base}$(SMD) & \textbf{85.07}&	\textbf{91.36}&	94.63&	\textbf{91.29}&	\textbf{88.91}&	\textbf{92.20}&	\textbf{76.51}&	\textbf{88.57}\\
        \midrule
        SRoBERTa$_{base}$ & 84.91&	90.83&	92.56&	88.75&	90.50&	88.60&	78.14&	87.76\\
        SimCSE-RoBERTa$_{base}$  & 84.92&	92.00&	94.11&	89.82&	91.27&	88.80&	75.65&	88.08\\
        TransAug-RoBERTa$_{base}$(SMD) & 85.08&	91.68&	94.61&	90.68&	91.32&	90.20&	76.46&	88.58\\
        SimCSE-RoBERTa$_{large}$ & \textbf{88.12}&	92.37&	95.11&	90.49&	\textbf{92.75}&	91.80&	76.64&	89.61\\
        TransAug-RoBERTa$_{large}$(SMD) & 87.22&	\textbf{92.66}&	\textbf{95.22}&	\textbf{91.34}&	92.59&	\textbf{93.40}&	\textbf{77.62}&	\textbf{90.01}\\
    \end{tabular}
    \caption{Performance on transfer tasks on the SentEval benchmark. All results are from ~\citealp{gao2021simcse,ni2021sentence,reimers2019sentence}. SMD represents the model is pre-trained on source-mixed dataset.}
    \label{tab:transfer}
\end{table*}

\subsection{Evaluation Results}

Following the same setting as previous works~\cite{gao2021simcse,ni2021sentence}, we evaluate using SentEval which includes 7 transfer and 7 STS tasks, the main goal of sentence
embeddings is to cluster semantically similar sentences, and take STS result as the main evaluation metric.

\subsubsection{Semantic textual similarity tasks}

We evaluate TransAug under zero-shot and fine-tuned settings. To fairly compare with previous works~\cite{gao2021simcse,ni2021sentence}, we adopt 7 STS tasks including STS 2012–2016~\cite{agirre2012sem,agirre2013sem,agirre2014semeval,agirre2015semeval,agirre2016semeval}, STS Benchmark~\cite{cer2017semeval} and SICK-Relatedness~\cite{marelli2014sick}. For STS, sentence embeddings
are evaluated by how well their cosine similarities correlate with human annotated similarity scores, which has been widely used in measuring the discriminative power of sentence embeddings. Suggested by~\citealp{gao2021simcse}, we also report Spearman’s correlation coefficients.

We start from pre-trained checkpoints of BERT or RoBERTa as backbone. For comprehensive comparison, we divide the comparison into 3 track: BERT track, RoBERTa track and State-of-the-art track. Specifically, BERT track includes Sentence-BERT~\cite{reimers2019sentence}, CT-BERT~\cite{carlsson2020semantic}, and SimBERT. RoBERTa track includes SimRoBERTa and Sentence-RoBERTa. In the term of State-of-the-art track, we compare with Sentence-T5~\cite{ni2021sentence} 11B model, which contains 11 billion parameters. Table \ref{tab:sts} reports the evaluation results on 7 STS tasks. TransAug can substantially improve results on all the datasets with or without extra NLI supervision, greatly outperforming the previous state-of-the-art models. 

Specifically, TransAug outperforms the averaged Spearman’s correlation of SimCSE by 0.89-2.65 under zero-shot setting. When using NLI datasets, TransAug-BERT$_{base}$ further pushes the state-of-the-art results from 84.94 to 85.15. The gains are more pronounced on RoBERTa encoders, and our TransAug achieves 85.60 with RoBERT$_{large}$.




\subsubsection{Transfer Tasks}

We evaluate on the following transfer tasks: MR~\cite{pang2005seeing}, CR~\cite{hu2004mining}, SUBJ~\cite{pang2004sentimental},
MPQA~\cite{wiebe2005annotating}, SST-2~\cite{socher2013recursive}, TREC~\cite{voorhees2000building} and MRPC~\cite{dolan2005automatically}. We employ the default configurations from SentEval\footnote{https://github.com/facebookresearch/SentEval}.
Results on transfer tasks are shown in Table \ref{tab:transfer}. 

Benefited from the large scale of parallel translation datasets that boost the power of contrastive learning, TransAug learns more generalized sentence representations than previous approaches, and improves performance on transfer tasks.



\subsection{Ablation Studies}


We investigate the impact of multilingual semantic distillation, the multilingual teacher-student architecture and different pooling methods. Our benchmark used in this section is the TransAug-BERT$_{base}$ (WMT) without any fine-tuning.

\subsubsection{Choices of Chinese Encoder}
\label{exp of msd}

In Section \ref{MSD}, we have briefly provided the reason why we do not use a pre-trained encoder in stage one. To further support our claim, in stage one, instead of distillation, we train two pre-trained Chinese sentence encoders. One is a RoBERTa$_{large}$ model trained with CCL, the other is a SimCSE-Roberta$_{large}$ model. Both are trained on Chinese NLI dataset\footnote{https://github.com/pluto-junzeng/CNSD}. In stage two, the pre-trained encoders and distilled encoder follow the same setting. We evaluate on SST-B development set and report the result in table \ref{EOM}. As shown, distilled model improves from 86.57 to 88.08 than pretrained model.

\renewcommand{\arraystretch}{1.2}
\begin{table}[!th]
  \centering
  \begin{tabular}{c||c|c|c}
    \hline
     Models & PT-SimCSE & PT-CCL & DT \\
    \hline
    STS-B  &86.06 &86.57  &\textbf{88.08}  \\
    \hline
  \end{tabular}
  \vspace{-2mm}
    \caption{\textbf{Comparison of distilled and pretrained encoders.} PT represents 'pre-trained' while DT represents 'distilled'. DT is TransAug-BERT$_{base}$ (WMT), PT-SimCSE and PT-CCL are RoBERTa$_{large}$ and SimRoberta$_{large}$ that trained with SimCSE and CCL strategies, respectively.}
  \label{EOM}
  \vspace{-2mm}
\end{table}




\subsubsection{Choices of Training Strategies}
\label{MTA}

In Section \ref{Preliminary}, we introduce two common strategies in machine translation approaches for handling translated sentence pairs. Figure \ref{fig:comparison} shows the difference between TransAug and these works. To show the effectiveness of our cross-lingual contrastive learning scheme, we train models with single multilingual encoder, regular dual
encoder and our TransAug architecture, respectively, and evaluate their performance on STS-B development set.

For dual encoder, we use the same distilled Chinese encoder from stage one and a BERT$_{base}$, then train via contrastive learning, instead of freezing the parameters of distilled Chinese encoder. In the term of single encoder, we adopt a RoBERTa$_{base}$-xlm~\cite{lample2019cross} model that accept multilingual input, and train this model following the same method as SimCSE for RoBERTa$_{base}$. Both are trained on WMT dataset.



\renewcommand{\arraystretch}{1.2}
\begin{table}[!th]
  \centering
  \begin{tabular}{c||c|c|c}
    \hline
     Models & DE & XLM & TBW \\
    \hline
    STS-B  &68.10 &72.71  &\textbf{88.08}  \\
    \hline
  \end{tabular}
  \vspace{-2mm}
    \caption{\textbf{Comparison of different strategies for translated sentence pairs.} DE, XLM and TBW represent dual encoder, single multilingual encoder and our TransAug-BERT$_{base}$(WMT).}
  \label{architecture}
  \vspace{-2mm}
\end{table}

The result of the correlation analyses is shown in \ref{architecture}. The multilingual  teacher-student architecture exhibits the best result, showing its great advantages for cross-lingual contrastive learning. 



\begin{figure}[t]
\centering
\includegraphics[scale=0.27]{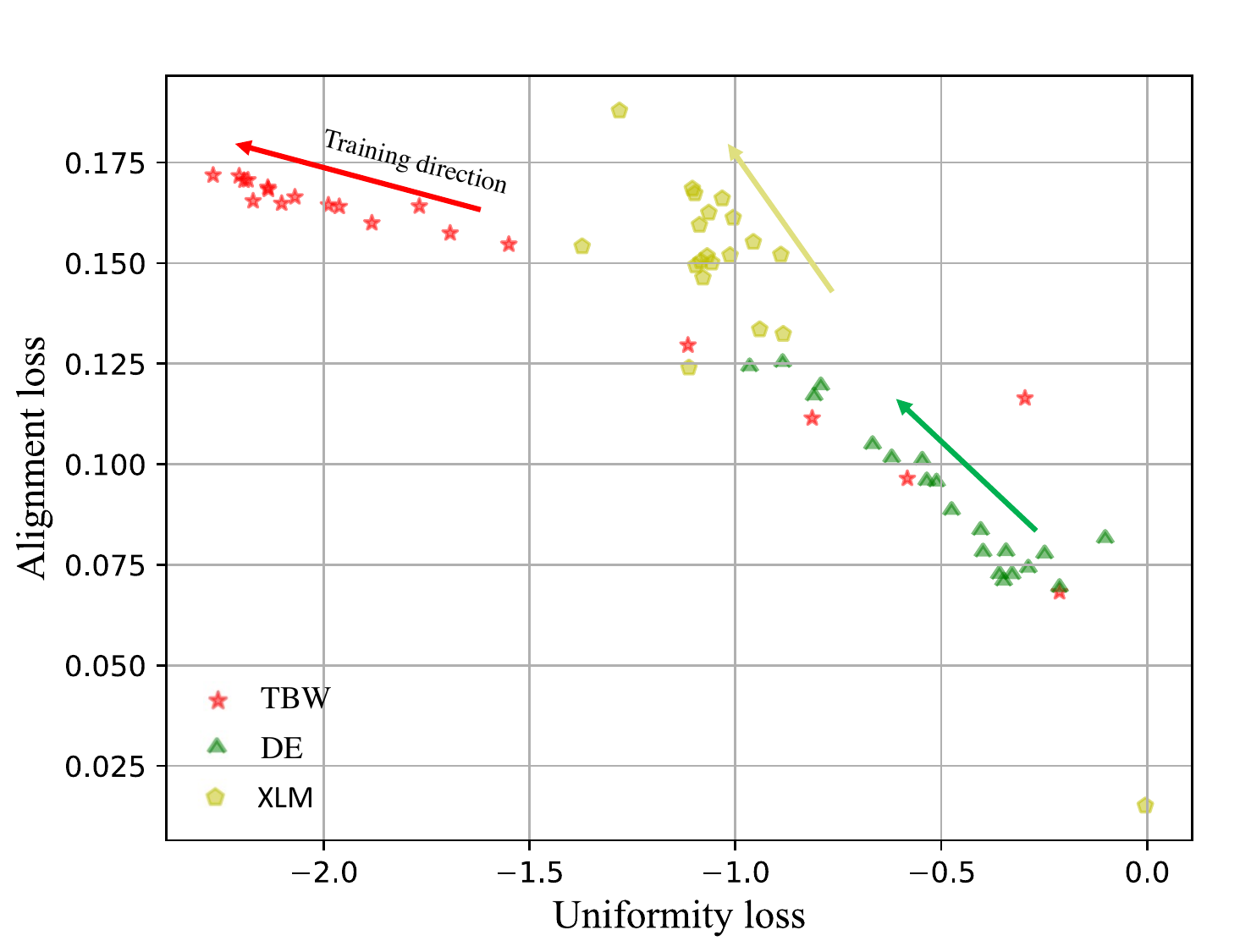}
\caption{Alignment and uniformity plot. We visualize checkpoints every 100 training steps, and the arrows indicate the training direction. For both `align' and `uniform', lower numbers are better. TBW, DE, XLM represent the TransAug-BERT$_{base}$(WMT), dual encoder architecture, and the RoBERTa$_{base}$-xlm model trained on WMT dataset, respectively.}
\label{fig:training direction}
\end{figure}

To further analyze its effectiveness, we evaluate the training process with alignment and uniformity~\cite{wang2020understanding} suggested by SimCSE~\cite{gao2021simcse}. We take the checkpoint every 100 steps during training and calculate the alignment and uniformity loss. As clearly shown in Figure \ref{fig:training direction}, all
models greatly improve uniformity, especially TransAug-BERT$_{base}$. However, the
alignment of the two counterparts degrades drastically, while our TransAug-BERT$_{base}$ keeps
a steady alignment thanks to a frozen operation.

\subsubsection{Pooling Methods}

As suggested by previous works~\cite{gao2021simcse}, pooling strategies make difference on the performance. Li et al~\cite{li-etal-2020-sentence} shows that taking the average embeddings of pre-trained model leads to better performance than [CLS]. Here, we consider three different pooling settings: (1) Average Pooling, (2) CLS, (3) CLS before pooler. Table \ref{pooling} shows the comparison between different pooling methods in TransAug. We evaluate on STS-B development set. As shown, we find that CLS before pooler method works the best for TransAug.



\renewcommand{\arraystretch}{1.2}
\begin{table}[!th]
  \centering
  \begin{tabular}{c||c|c|c}
    \hline
     Models & CLS & AVG & CBP \\
    \hline
    STS-B  &85.19 &87.28  &\textbf{88.08}  \\
    \hline
  \end{tabular}
  \vspace{-2mm}
    \caption{\textbf{Performance of different pooling methods.} CBP represent [CLS] before pooler method.}
  \label{pooling}
  \vspace{-2mm}
\end{table}

\section{Conclusion}
In this work, we propose TransAug, a simple but effective data augmentation method for sentence embeddings via translation. To utilize the translated pairs, we introduce a two-stage paradigm to advances the state-of-the-art sentence embeddings. We demonstrated that TransAug achieves a new state-of-art on both downstream transfer tasks and standard semantic textual similarity (STS), outperforming both SimCSE and Sentence-T5.

\bibliography{acl}
\bibliographystyle{acl}

\appendix

\end{document}